\def\year{2021}\relax
\documentclass[letterpaper]{article} 
\usepackage{aaai21}  
\usepackage{times}  
\usepackage{helvet} 
\usepackage{courier}  
\usepackage[hyphens]{url}  
\usepackage{graphicx} 
\usepackage{natbib}  
\usepackage{caption} 
\title{ Second Order Optimization for Adversarial Robustness and Interpretability }
\author{
    Theodoros Tsiligkaridis,\textsuperscript{\rm 1 \textdagger \textasteriskcentered}
    Jay Roberts \textsuperscript{\rm 1 \textdaggerdbl }
    \footnote{Authors have made equal contributions.}\\
}
\affiliations{
    \textsuperscript{\rm 1} Massachusetts Institute of Technology Lincoln Laboratory\\
    \textsuperscript{\textdagger} Artificial Intelligence Technology Group, \textsuperscript{\textdaggerdbl} Homeland Sensors and Analytics Group \\
    244 Wood St, Lexington, MA 02421\\
    \{ttsili, jay.roberts\}@ll.mit.edu
}

\usepackage[switch]{lineno}  %
\usepackage{natbib}
\usepackage{amsmath,amssymb,mathtools,amsthm,bm,etoolbox}
\usepackage[linesnumbered,ruled]{algorithm2e}
\usepackage{subfig}
\usepackage{xcolor}

\newtheorem{theorem}{Theorem}

\newtheorem{prop}{Proposition}

\newcommand{\sgn}{\text{sgn}}
\newcommand{\grad}{\nabla}

\newcommand{\hessx}{\nabla^2_x}
\newcommand{\inp}[2]{\langle #1, #2 \rangle}

\newcommand{\EE}{\mathbb{E}}
\newcommand{\gradx}{\grad_{x}}
\newcommand{\gradv}{\grad_{v}}

\newcommand{\gradbld}{\text{\bf grad}}

\newcommand{\half}{\frac{1}{2}}

\DeclareMathOperator*{\argmin}{argmin}
\DeclareMathOperator*{\argmax}{argmax}

\newcommand{\nn}{\parallel}

\newcommand{\loss}{\ell}

\newcommand{\quadd}{\quad \quad}
\newcommand{\eps}{\epsilon}

\nocopyright
\begin{document}
\maketitle

\begin{abstract}
Deep neural networks are easily fooled by small perturbations known as adversarial attacks. Adversarial Training (AT) is a technique aimed at learning features robust to such attacks and is widely regarded as a very effective defense. However, the computational cost of such training can be prohibitive as the network size and input dimensions grow. Inspired by the relationship between robustness and curvature, we propose a novel regularizer which incorporates first and second order information via a quadratic approximation to the adversarial loss. The worst case quadratic loss is approximated via an iterative scheme. It is shown that using only a single iteration in our regularizer achieves stronger robustness than prior gradient and curvature regularization schemes, avoids gradient obfuscation, and, with additional iterations, achieves strong robustness with significantly lower training time than AT. Further, it retains the interesting facet of AT that networks learn features which are well-aligned with human perception. We demonstrate experimentally that our method produces higher quality human-interpretable features than other geometric regularization techniques. These robust features are then used to provide human-friendly explanations to model predictions.
\end{abstract}

\section{Introduction}

Deep neural networks (DNN) are powerful models that have achieved excellent performance across various domains \cite{LeCun:2015} by exploiting hierarchical representations of data. As these models are being deployed across industries, such as healthcare and autonomous driving, robustness and interpretability concerns become increasingly important. Several organizations have also have identified important principles of
artificial intelligence (AI) that include notions of reliability and transparency \cite{google:blog, microsoft:blog, dod:blog}.

One issues of such large capacity models is that small, carefully chosen input perturbations, known as adversarial perturbations, can lead to incorrect predictions~\cite{goodfellow:2015}. Various enhancement methods have been proposed to defend against adversarial perturbations \cite{kurakin:2017, ros2018:gradreg, madry2018:at, md2019:cure}. One of the best performing algorithms is adversarial training (AT)
\cite{madry2018:at}, which defends against strong adversarial perturbations by attacking the model during training. Computing these adversarial perturbations at each step of AT requires many iterations of a gradient-based optimization to be performed for each new minibatch. This
becomes computationally prohibitive as the model size and input dimensions grow. 
Training against weaker attacks can reduce this cost but leads to strong robustness against weak attacks and brittleness against stronger attacks. 
This can be due to gradient obfuscation \cite{Uesato:2018, Athalye:2018}, a phenomena where networks learn to defend against gradient-based attacks by 
making the loss landscape highly non-linear.
Another sign of gradient obfuscation is when adversarial attacks computed with a few iterations fail but black-box attacks successfully find adversarial perturbations \cite{Uesato:2018, simba:2019}.

Another major concern is interpretability of DNN decisions and explanation methods for AI system users or stakeholders. Insights into model behavior based on counterfactual explanations has the potential to be be very useful for users \cite{Wachter:2018}. However, standard networks do not have interpretable saliency maps and adversarial attacks tend to be visually imperceptible. Some popular explanation methods include layerwise relevance propagation (LRP) \cite{LRP:2015}, locally interpretable model-agnostic explanations \cite{LIME:2016}, and contrastive explanations \cite{Dhurandhar:2019}, but these methods only yield feature relevance and are susceptible to spurious correlations prevalent in standard networks \cite{Ilyas:2019}. The relationship between adversarial robustness and saliency map interpretability was recently studied in \cite{Etmann:2019} but experiments were based on gradient regularization. Furthermore, recent works \cite{Tsipras2018:tradeoff, Ilyas:2019} claim that existence of adversarial examples are due to standard training methods that rely on highly predictive but non-robust features, and make connections between robustness and explainability.

In this paper, we propose a quadratic-approximation of adversarial attacks that we incorporate into a regularizer which smooths the loss landscape and yields adversarial robustness. This smoothness implies that, in a neighborhood of
a point $x$, the loss landscape is well approximated by gradient and curvature information. Our models avoid gradient obfuscation by exploiting this information.
We refer to this second-order robust optimization approach as SCORPIO. We empirically show that networks trained with the SCORPIO regularizer combat gradient obfuscation and retain a high level of robustness against various types of strong attacks. Furthermore, we show how these networks improve interpretability and can be used to explain DNN predictions using a framework inspired by \cite{Dhurandhar:2019, Ilyas:2019}. Our main contributions are summarized below:

\begin{itemize}
\item A new regularizer is proposed (that may be adapted for any $L^p$ norm) with minimal tuning parameters based on projection-free Frank-Wolfe iterations on a local quadratic-approximation to the adversarial risk that incorporates second-order information.

\item It is shown that the
SCORPIO regularizer trains faster than AT, provides robustness nearly on the same level as AT, and outperforms prior gradient and curvature regularizers when evaluated under various types of strong white-box attacks.

\item It is demonstrated experimentally that SCORPIO better protects models against some black-box attacks than AT, suggesting that SCORPIO
suffers from less gradient obfuscation than AT.

\item It is shown that the quadratic-approximate attacks are still successful against robust models.

\item It is shown that the quality of saliency maps and adversarial perturbations improves for the SCORPIO regularizer.

\end{itemize}

\section{Background and Previous Work}
Consider $(x_i,y_i)\sim \mathcal{D}$ pairs of data examples drawn from distribution $\mathcal{D}$. The labels are assumed to span $K$ classes. 
The neural network function $f_\theta(\cdot)$ maps input features into logits, where $\theta$ are the  model parameters. The class probability scores are obtained using the softmax transformation $p_k(x)=e^{f_{\theta,k}(x)}/\sum_l{e^{f_{\theta,l}(x)}}$. The predicted class label is given by $\hat{y}(x)=\arg \max_k f_{\theta,k}(x)$.

The prevalent way of training classifiers is through empirical risk minimization (ERM):
\begin{equation*}
    \min_\theta \EE_{(x,y)\sim \mathcal{D}} [\loss(x,y;\theta)]
\end{equation*}
where the loss is the cross-entropy loss function given by $\loss(f_\theta(x),y)=\loss(x,y;\theta)=-y^T\log(p_\theta(x))$, where $y$ denotes the one-hot label vector.

Adversarial robustness for a classifier $f_\theta$ is defined with respect to a metric, here chosen as the $L^p$ metric associated with the ball $B_p(\epsilon)=\{\delta:\nn \delta \nn_p\leq \epsilon\}$, as follows. A network is said to be robust to adversarial perturbations of size $\epsilon$ at a given input example $x$ iff $\hat{y}(x)=\hat{y}(x+\delta)$ for all $\delta \in B_p(\epsilon)$, i.e., if the predicted label does not change for all perturbations of size up to $\epsilon$. The $\epsilon$ is often referred to as the
strength or budget of the attack.

Training neural networks using the ERM principle gives high accuracy on test sets, but leaves the network vulnerable to adversarial attacks. 
One of the most effective defenses against such attacks is adversarial training (AT) \cite{madry2018:at} which aims to minimize the adversarial risk instead,
\begin{equation} 
\label{eq:adv_risk}
    \min_\theta \EE_{(x,y)\sim \mathcal{D}} 
        \left[ 
            \max_{\delta \in B_p(\epsilon)}  \loss(x+\delta,y;\theta) \right]
            .
\end{equation}
The training procedure constructs adversarial attacks at given inputs $x$ that aim to solve the inner maximization problem. Common maximization
methods typically use a fixed number of gradient-ascent optimizations. One such method is projected gradient descent (PGD) that performs the iterative updates:
\begin{equation}
    \delta^{(k+1)} = P_{B_p(\epsilon)}( \delta^{(k)} + \alpha \grad_\delta \loss(x+\delta,y;\theta) )
\end{equation}
where $P_{B_p(\epsilon)}(z)=\arg \min_{u\in B_p(\epsilon)} \nn z - u \nn_2^2$ denotes the orthogonal projection onto the constraint set. The sign of the gradient has also been shown to be an effective perturbation. The computational cost of this method is dominated by the number of steps used to approximate the inner maximization, since an $N$ step PGD approximation to
the maximization (denoted PGD($N$)) involves $N$
forward-backward propagations through the network. While using fewer PGD steps can lower this cost, these amount to
weaker attacks which can lead to gradient obfuscation \cite{Papernot:2017, Uesato:2018}.

Prior works on regularization for adversarial robustness include gradient \cite{Lyu:2015, ros2018:gradreg} and curvature based penalties \cite{md2019:cure}. Gradient methods have not been shown to yield good robustness against strong attacks, while the relationship between small curvature and high robustness has been better established, but still leaves significant room for improvement. 
Based on the connection between curvature and robustness, a regularizer that promotes local linearity near training examples was proposed in \cite{qin:2019} which is based on minimizing an upper bound on the adversarial risk via several PGD steps. These methods argue that flattening the decision boundary via the loss is a suitable defense against adversarial attacks.

\section{Approximating the Adversarial Attack}

%
%
For this section we suppress the explicit dependence 
on the network and target and let
$\loss(x) := \loss(f_\theta(x), y)$.
Rather than approximate the maximization of 
\eqref{eq:adv_risk} directly
first define quadratic-approximate
risk function
\begin{equation}
    \label{eq:qaud risk def}
    Q(v; x) :=
        \gradx \loss(x) \cdot v
        +
        \half
        v^T
        \hessx \loss(x)
        v
\end{equation}
and then propose solving the quadratic approximate risk optimization problem,
\begin{equation}
\label{eq:quad sub max}
 v_Q = 
    \argmax_{
        |v|_p \leq \epsilon
        } \ Q(v; x)
        .
\end{equation}
Intuitively, we expect this quadratic approximate loss,
$\loss(x + v_Q)$ to play the
role of the adversarial risk, and $v_Q$ to play that of
an attack.

In order to compare the quadratic-approximate attack and the adversarial
attack we let
\begin{equation*}
    v_A = \argmax_{
                v \in B_p(\epsilon)
                } \loss(x + v)
\end{equation*}
be the optimal adversarial attack and define $L_A = \loss(x + v_A)$ and $L_Q = \loss(x + v_Q)$ to be the corresponding
worst case adversarial-loss and the worst case quadratic-approximate loss,
respectively.
The following theorem gives bounds on these two losses
in terms
of the regularity of the loss, and so implicitly the network.

\begin{theorem}
\label{thm:adv loss gap}
    Let $L_A$ and $L_Q$ be the worst case adversarial loss and worst
    case quadratic-approximate loss at some $x$ over $B_p(\epsilon)$.
    If $\loss \circ f_\theta \in C^3(B_p(\epsilon))$ then
    \begin{equation}
        \label{eq:loss diff}
        |L_A - L_Q|
            \leq \ 
                2 \epsilon \| \gradx \loss(x) \|_q
                +
                2
                {\epsilon^2} 
                \| \hessx \loss(x) \|_{p,q}
                +
                \frac{\epsilon^3 M}{3}
    \end{equation}
    where $M$ is the supremum-norm 
    of all derivatives of $\loss(x)$ of order three,
    over $B_p(\epsilon)$.
\end{theorem}

It is important here to emphasize the implications of Theorem \ref{thm:adv loss gap}.
The amount by
which the quadratic-approximate attack is {\bf either stronger or weaker} than
a full adversarial attack is controlled by the strength of the attack
and by the regularity of the network. 
The advantage of using the quadratic-approximate
attack is that the optimization \eqref{eq:quad sub max}
can be well approximated with less computation complexity than
standard AT.


For $L^2$ attacks larger attack strengths, $\epsilon$, can still result
in imperceptible changes to the input. However, $L^\infty$ attacks
are considered stronger attacks since they can deceive a network
using a smaller strength perturbation, and so modify the input in a less perceptible
way. 
Theorem \ref{thm:adv loss gap} suggests that the gap between the
two attacks should be smaller in the $L^\infty$ case than $L^2$. This
is demonstrated experimentally below.

It has been shown experimentally that many robust models,
including AT and the geometric-regularization methods mentioned
above, have the more regular loss landscapes than their
non-robust counterparts \cite{Lyu:2015, ros2018:gradreg, md2019:cure, qin:2019}. Theorem \ref{thm:adv loss gap} implies that greater
regularity of the network, and thus its decision boundary,
will result in a smaller gap between the quadratic approximate
risk and the adversarial risk. This suggests that the
quadratic approximate attacks, $v_Q$, should fool robust models at
similar rates to PGD attacks. However,
standard non-robust models
can be much less regular, 
suggesting quadratic-approximate attacks will not perform as well as
min-max attacks. We demonstrate this phenomena experimentally below.

The
new approximate attack into the loss via a regularizer that penalizes
deviation from the original prediction
\begin{equation}
    \loss_Q = \loss(x + v_Q) - \loss(x).
\end{equation}
Since $\loss_Q = Q(v_Q) + R_{x,3}(v_Q)$, adding it as a regularizer incorporates both gradient and curvature terms not explicitly exploited by AT. Further it does this without directly computing these derivatives, as done in gradient and curvature regularizations, which can lead to expensive backprop steps during training. 

\subsection{Finding the Optimal Approximate Attack}

Let $v^k$ be the approximate
solution to \eqref{eq:quad sub max} at step $k$, $g = \gradx \loss(x)$, and $H = \hessx \loss(x)$. 
A good initialization for the iteration is
$v^0 = \epsilon g / \nn g \nn_p$, since this maximizes the
inner product term; moreover,
it has been
demonstrated experimentally that the gradient direction
is well aligned with the direction of maximum curvature of
neural networks \cite{Jetley:2018, Fawzi:2018} so $v^0$ also serves as a strong initialization for
the curvature term $v^T H v$.

An obvious first choice would
be projected gradient decent as is done in AT, but
instead of optimizing the adversarial risk we would
optimize \eqref{eq:qaud risk def}. However,
this method involves tuning the step size $\alpha$,
and we experimentally found that it provided less robustness than the next approach for the same number of steps. 
%
%
%
%
%
%

For these reasons we use a
Frank-Wolfe (FW) iteration, ~\cite{jaggi13}. This method
first solves a linearized version of the problem over convex
sets,
\begin{equation}
    \label{eq:FW basic}
    \begin{cases}
        \displaystyle s^{k} := \argmax_{
                    \nn s \nn_p \leq \epsilon
                    } \ 
                 {s \cdot
                    \gradv Q(v^k)
                    }
            \\
        v^{k+1} := (1 - \gamma^k)v^k + \gamma^k s^{k}
    \end{cases}
\end{equation}
where $\gamma^k = 2/(2+k)$ is the step size. The FW sub-problem can be solved exactly for any $L^p$ and
the optimal $s^k = P_{FW}(v^k;p)$ is given by
\begin{equation}
    \label{eq:lp opt s}
    P_{FW}(v^k;p) = \alpha \cdot \sgn(\gradv Q(v^k)_i) \ |\gradv Q(v^k)_i|^{p/q}
\end{equation}
with $\alpha$ chosen so that $\nn s^k \nn_p = \epsilon$.

Note FW does not require a projection onto the $L^p$ ball
which is non-trivial for $p$ not in $\{1,\ 2,\ \infty \}$. Another advantage of FW 
is the $\gamma^k$ do not need
to be tuned. 
%
Interestingly,
for the special case of $L^\infty$ attacks the optimal solution becomes the signed gradient of the quadratic approximation, i.e. $s^k = \epsilon \ \sgn(\gradv Q(v^k) )$. This suggests a connection to the Fast Gradient Sign Method (FGSM) of adversarial attacks, ~\cite{goodfellow:2015}.

\begin{prop}
\label{lem:fw ascend}
    If $v^0 = g$ and $H$ is positive-semi-definite then the sequence of $v^k$ defined by \eqref{eq:FW basic} is a non-decreasing sequence for $Q$. That is for all $k\geq0$ 
    \begin{equation*}
        Q(v^{k}) \geq Q(v^{k-1}) \geq \hdots \geq 0
        .
    \end{equation*}
\end{prop}

The condition that the Hessian of the loss, $H$, be positive semi-definite
has been shown to hold locally for all $x$, excluding a set of measure 0, 
when the network uses ReLU activations and the loss is categorical cross
entropy \cite{singla2019understanding}. 

\subsection{Computational Complexity}
At each iteration of FW the gradient of the
quadratic approximate risk are required. The term
\begin{equation}
\label{eq:quad grad}
    \gradv Q(v) = \gradx \loss(x) + \hessx \loss(x) v
\end{equation}
requires a Hessian vector product. We note that use of this gradient into (\ref{eq:FW basic}) incorporates first- and second-order loss information into each iteration. 
One can
compute the Hessian vector product  
in 1 forward and 2 backward passes. However, 
computing backward passes
during training is expensive. Further, it was
noted in \cite{md2019:cure} that using highly
localized curvature information conferred less
robustness to the model and using a finite difference
with relatively large step sizes allowed the
network to regularize the loss curvature in a neighborhood
of the sample point. For these reasons 
we propose using two
approximations to the Hessian vector product. First,
the Forward Euler (FE) approximation
\begin{align}
    \label{eq:foeuler}
    \hessx \loss(x) v
        \ \approx \ 
            \frac{
            \left[
                \gradx \loss(x + h v)
                -
                \gradx \loss(x)
            \right]}{h}
\end{align}
and the Central Difference (CD) approximation
\begin{align}
    \label{eq:centdiff}
    \hessx \loss(x) v
    \ \approx \ 
            \frac{
            \left[
                \gradx \loss(x + h v)
                -
                \gradx \loss(x -h v)
            \right]}{2h}
            .
\end{align}
%

Naively, computing \eqref{eq:quad grad} with either
finite difference costs three forward and one
backward pass; however, the gradient at $x$ can be computed
once and in the case of FE it can be recycled to be
used into the finite-difference term. The cost of
 an $N$-iteration FW is then $N+1$ forward and backward
passes for FE and $2N + 1$ for CD. The FW-FE algorithm with
gradient recycling is given in Algorithm \ref{alg:fw-fe}.

In our experiments we show that even for 
$N=3$ both FE- and CD-SCORPIO, costing 4 and 7 forward-backward
passes respectively, achieve nearly the same robustness
as PGD(10)-AT which costs 10 forward-backward passes.

Note that if sufficient memory is
available CD can be further optimized by parallelizing the computation
of the forward pass of $x\pm hv$ 
allowing modern
deep learning frame-works on GPU hardware to parallelize
the majority of the computations. In practice on CIFAR-10 and
SVHN with batch sizes of 512 we see little difference in the 
FE and CD training times.

\begin{algorithm}
    \SetKwInOut{Input}{Input}
    \SetKwInOut{Output}{Output}
    
    \Input{$N$,$\epsilon$, $h$, p}
    \Output{Approximate solution, $v^*$, to \eqref{eq:quad sub max}.}
    
    $l^0$ = $\loss(f_\theta(x_0),\  y_0)$
    
    $g^0 = \gradbld [l^0, x_0]$
    
    $v^* \gets g^0$
    
    \For{ $k \in \{0, ..., N-1\}$}{
        
        $\gamma^k = 2/(2 + k)$
        
        $l^k$ =
            $\loss(
                f_\theta(x_0 + h v^*),\  y_0$
                )
                
        $g^k = \gradbld [ l^k, x_0]$

        
        $s^k = \epsilon * P_{FW}(
                g^0 + \frac{1}{h}(g^k - g^0)
                ; p$
            )

        $v^* \gets (1- \gamma^k) v^* + \gamma^k s^k$
        }
    
    \caption{Forward Euler Frank Wolfe (FE-FW) applied to a neural network $f_\theta$, at an input $x_0$ with corresponding target $y_0$ and a loss $\loss$.}
    \label{alg:fw-fe}
\end{algorithm}

\section{Experimental Results}

We evaluate the performance of our proposed SCORPIO regularizer against standard networks, gradient regularizered networks (Grad-Reg) \cite{ros2018:gradreg}, curvature regularized networks (CURE) \cite{md2019:cure}, and adversarial training (AT) \cite{madry2018:at}.

\subsection{Implementation}
The models, training, and evaluation routines for our experiments were performed 
using the robustness library from
~\cite{robustness} which we modified to to support the SVHN dataset. Experiments were run on 2 Volta V100
GPUs using the computing resources at ({\bf Computing credits censored to maintain anonymity}).

Hyperparameters were tuned for each experiment 
with the goal of achieving near the clean accuracy of AT. For SCORPIO models, the finite difference step size $h$
and regularization strength $r$ were tuned via a manual search for both between 0.5 and 1.5. 
For tuning CURE, grad-reg, we chose the regularization strength to be 
as large as possible 
while still nearly matching the clean accuracy of AT. For AT we ran PGD with $10$ steps and chose the best model that achieves the highest adversarial accuracy at $\eps=0.5$ for $L^2$ robustness and $\eps=8/255.0$ for $L^\infty$ robustness.
Specifics can be found in the supplementary materials.

\subsection{Adversarial Robustness}
To test the robustness of our network, we consider a variety of adversarial attacks, including untargeted and targeted attacks towards a random class $r\sim Unif(\{1,\dots,K\}\backslash y)$. The classification margin is defined as:
\begin{equation} \label{eq:margin}
   M(x,y) 
        = \log p_y(x) - \max_{j\neq y}\log p_{j}(x)
\end{equation}
The following loss functions are used to obtain low adversarial accuracy:

{\bf (UL) Untargeted-loss: }
$\max_{\delta \in B(\epsilon)} l(x+\delta,y)$

{\bf (TL) Rand.Targeted-loss: }
$\max_{\delta \in B(\epsilon)} l(x+\delta,r)$

{\bf (UM) Untargeted-margin:  }
$\min_{\delta \in B(\epsilon)} M(x+\delta,y)$

{\bf (TM) Rand. Targeted-margin: }
$ \max_{\delta \in B(\epsilon)} M(x+\delta,r)$

The performance metric used is the accuracy on the test set after the attack is applied, i.e., adversarial accuracy.
We empirically observed that the margin-based attacks on robust models are stronger than loss-based attacks for robust models.

\begin{figure}[!tbp]
  \centering
  \subfloat[CIFAR10 $L^2$ robustness.]{\includegraphics[width=0.23\textwidth]{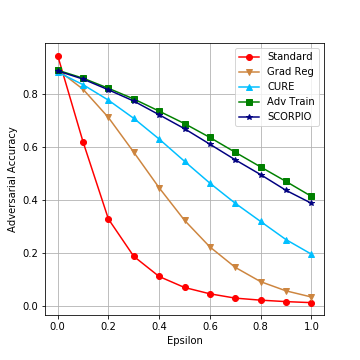}\label{fig:evalPGDL2}}
  \hfill
  \subfloat[CIFAR10 $L^\infty$ robustness.]{\includegraphics[width=0.23\textwidth]{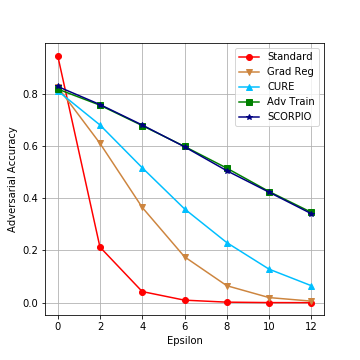}\label{fig:evalPGDLinf}}
  \newline
  \subfloat[SVHN $L^2$ robustness.]{\includegraphics[width=0.23\textwidth]{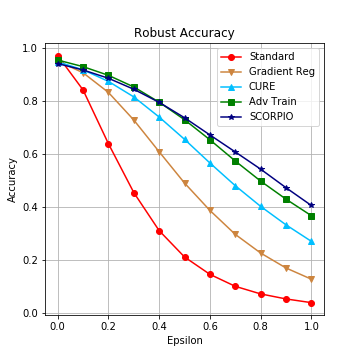}\label{fig:svhnevalPGDL2}}
  \hfill
  \subfloat[SVHN $L^\infty$ robustness.]{\includegraphics[width=0.23\textwidth]{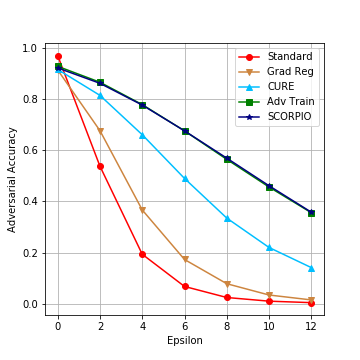}\label{fig:svhnevalPGDLinf}}
  \caption{Adversarial accuracy as a function of $\epsilon$ on CIFAR-10 (a,b) and SVHN (c,d) test set. Adversarial attacks on the loss using PGD(10) were used for evaluating robustness. We observe SCORPIO-FE(3) achieves nearly the same robustness levels as adversarial training with PGD(10) adversary and outperforms gradient and curvature regularizers by a large margin. }
\end{figure}

On CIFAR-10 SCORPIO outperforms both Grad-Reg and CURE by 
a sizable margin and achieves robustness close to that of
AT for increasingly strong attacks. The gap between SCORPIO and
AT is smaller for the $L^\infty$ case than the $L^2$, see
Figure \ref{fig:evalPGDL2} and \ref{fig:evalPGDLinf}. In Table \ref{tab:L2 cifar attacks} and \ref{tab:Linf cifar attacks} we report results for the
specific attacks in the $L^2$ and $L^\infty$ case, respectively. 
In general untargeted
attacks are stronger than targeted and margin attacks are stronger
then loss based ones.
For all attacks the trend holds that SCORPIO outperforms the
geometric regularizers and is close to AT. 

The SVHN dataset 
followed the same trends. In all cases SCORPIO outperforms Grad-Reg
and CURE and is close to, sometimes better than, AT.
see Figure \ref{fig:svhnevalPGDL2} and \ref{fig:svhnevalPGDLinf} and
supplemental tables \ref{tab:L2 svhn attacks} and \ref{tab:Linf svhn attacks} for details.

\begin{figure}
    \centering
    \subfloat[$L^2$ adversarial PGD(10) attacks with loss / margin at $\epsilon=0.5$]{
    \begin{tabular}{l|ccccc}
    Method       & Clean & UL & TL & UM & TM  \\
    \hline
    Standard     & 94.5 & 6.9  & 27.3 & 6.6  & 29.4 \\
    Gradient Reg & 89.2 & 32.3 & 70.2 & 32.2 & 66.9 \\
    CURE         & 88.3 & 54.6 & 81.5 & 53.7 & 79.5 \\
    SCORPIO (FE,N=3) & 88.7 & 66.9 & 85.3 & 66.8 & 83.8 \\
    SCORPIO (CD,N=3) & 89.9 & 66.4 & 85.4 & 66.2 & 83.8 \\
    Adv Train (PGD-10)    & 89.0 & 68.8 & 85.4 & 68.8 & 84.6 \\
  \end{tabular}
  \label{tab:L2 cifar attacks}
    }
    \hfill
    \subfloat[$L^\infty$ adversarial PGD(10) attacks with loss / margin at $\epsilon=8/255$]{
      \begin{tabular}{l|ccccc}
    Method       & Clean & UL & TL & UM & TM  \\
    \hline
    Standard     & 94.5 & 0.2  & 13.2 & 0.3 & 14.4 \\
    Gradient Reg & 82.5 & 6.5  & 40.0 & 7.1 & 34.8 \\
    CURE         & 81.1 & 23.0 & 60.7 & 22.4 & 54.3 \\
    SCORPIO (FE,N=3) & 82.7 & 50.5 & 75.2 & 48.7 & 73.0 \\
    SCORPIO (CD,N=3) & 83.3 & 49.2 & 74.6 & 47.2 & 71.9 \\
    Adv Train (PGD-10)    & 81.7 & 51.5 & 75.1 & 49.9 & 72.7 \\
  \end{tabular}
  \label{tab:Linf cifar attacks}
    }
    \caption{Model accuracy on CIFAR-10 test set against various attacks. Our proposed regularizer, SCORPIO, outperforms prior gradient and curvature regularization methods achieves nearly the same level of robustness as AT in all cases.}
    \label{fig:my_label}
\end{figure}

\begin{table*}[t]
  \centering
  \begin{tabular}{l|ccccc}
    Method       & Clean & PGD(10) & PGD(20) & PGD(40) & PGD(100)  \\
    \hline
    SCORPIO (FE,N=1)   & \textbf{89.42} & \textbf{66.35} & \textbf{66.06} & \textbf{65.98} & \textbf{65.97} \\
    SCORPIO (CD,N=1)   & \textbf{88.16} & \textbf{66.03} & \textbf{65.83} & \textbf{65.74} & \textbf{65.72} \\
    Adv Train (PGD-3) & 87.62 & 63.19 & 62.89 & 62.81 & 62.81 \\
  \end{tabular}
  \caption{Model accuracy on CIFAR-10 test set for untargeted marging (UM) based adversarial $L^2$ PGD attacks on the margin at $\epsilon=0.5$ for robust networks trained against $L^2$ adversary. The step size for PGD was set to $\alpha=0.1$. Both methods are trained starting from pretrained standard networks with fine-tuning. Our proposed regularizer, SCORPIO, with one FW step $N=1$, outperforms the PGD-based adversarial training with $N=3$ steps. }
  \label{tab:FW1vsPGD3-L2}
\end{table*}

\subsection{Approximate Quadratic Attacks}

For these experiments we compare $L^2$ PGD adversarial attacks on the loss against the quadratic-approximate attacks obtained by approximating solutions to \eqref{eq:quad sub max}. We compute adversarial attacks using $N=10$ PGD steps and approximate the quadratic-approximate attacks with $N=3$ steps of Frank-Wolfe (FE). We see for models with smoother loss surfaces, AT, SCORPIO, and CURE, the quadratic-approximate attack performs as well or better than traditional adversarial attacks. However, for standard models PGD attacks are superior than the quadratic-approximate attack. Based on Theorem \ref{thm:adv loss gap} we believe this is a demonstration of the quadratic-approximate attack's ability to exploit regularized decision boundaries which are not present in standard models. 

We emphasize that although quadratic-approximate attacks are not as strong as PGD against standard models, training standard models with the proposed SCORPIO regularizer none-the-less confers strong robustness as shown above.

\begin{figure}[!tbp]
  \centering
  \subfloat[CIFAR-10.]{\includegraphics[width=0.23\textwidth]{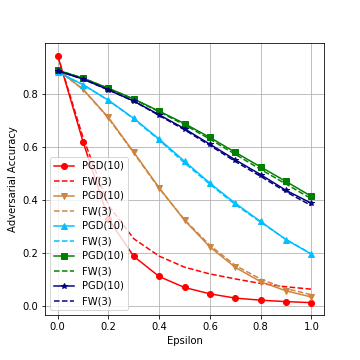}}
  \hfill
  \subfloat[SVHN.]{\includegraphics[width=0.23\textwidth]{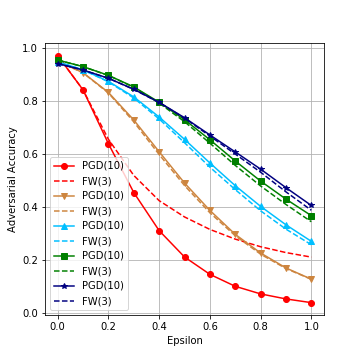}}
  \caption{Comparison between adversarial risk attacks (solid lines) and the quadratic-approximate risk attacks (dashed lines) on CIFAR-10 (right) and SVHN (left) datasets for varying strength $L^2$ attacks. Quadratic approximate attacks are slightly more successful than adversarial attacks on robust models (green, purple, blue, orange) but much less successful against standard models (red) likely due to these models having less regular decision boundaries.}
    \label{fig:quad attack}
\end{figure}


\subsection{Evaluating Gradient Masking}

\begin{figure}[!tbp]
  \centering
  \subfloat[Adversarial Training.]{\includegraphics[width=0.23\textwidth]{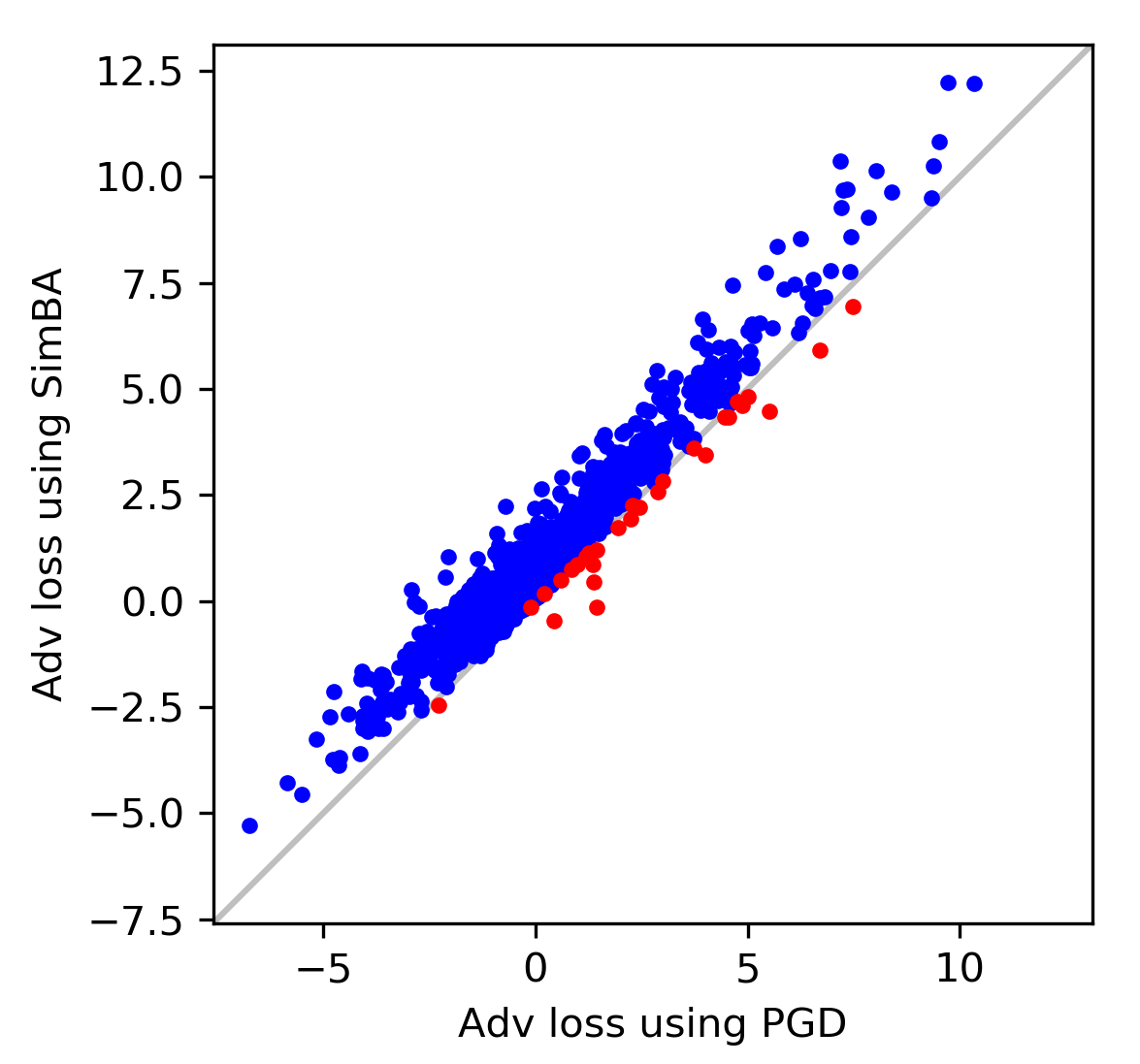}\label{fig:gradmaskat}}
  \hfill
  \subfloat[SCORPIO.]{\includegraphics[width=0.23\textwidth]{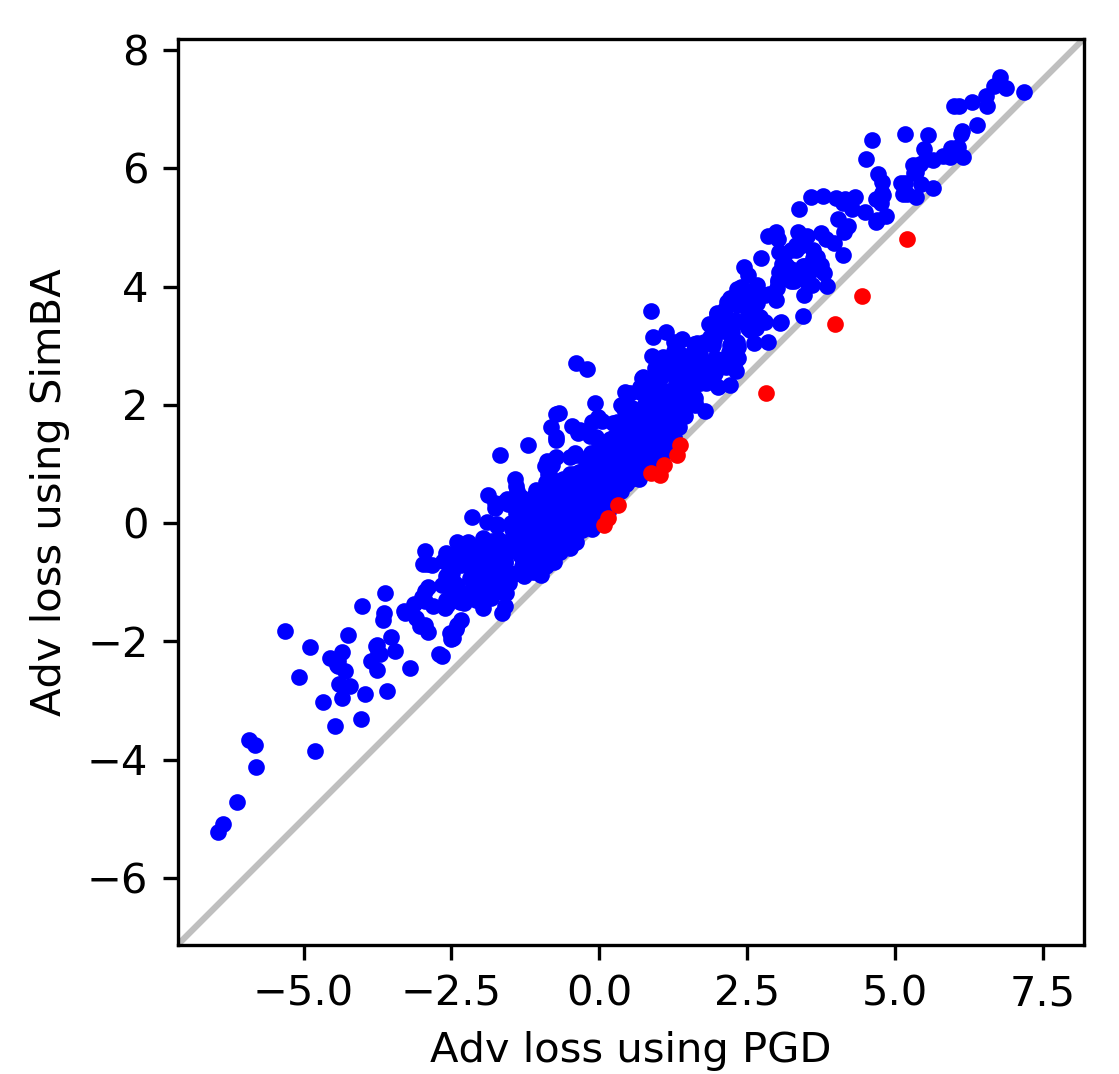}\label{fig:gradmaskscorpio}}
  \caption{Gradient masking analysis for network trained with adversarial training PGD(10) and SCORPIO-$L^\infty$. Adversarial loss here refers to the logit difference/margin (\ref{eq:margin}) and was computed with SimBA ($T=200$, $\epsilon=0.2$) for the y-axis and PGD(100) at $\epsilon=8/255$ for x-axis on a set of $1000$ test points. Points near the line $y=x$ indicate both types of attacks found similar adversarial perturbations, while points below the line shown in red imply that SimBA identified stronger attacks than PGD. SCORPIO exhibits a higher resistance to gradient masking. }
\end{figure}

We evaluate gradient masking by following a similar evaluation protocol as in \cite{md2019:cure} inspired by \cite{Uesato:2018}. Table \ref{tab:FW1vsPGD3-L2} shows that our method achieves similar adversarial accuracy on the test set when evaluated against stronger PGD attacks. Thus increasing the attack's complexity does not deteriorate the adversarial accuracy significantly. Furthermore, we evaluate our model against black box gradient-free method SimBA \cite{simba:2019}, and compared the margin computed using SimBA and PGD for a large batch of test points in Figure \ref{fig:gradmaskat} and \ref{fig:gradmaskscorpio} and observe that both methods lead to a similar adversarial loss except on a very small subset of points (12 out of 1000) for SCORPIO which improves upon the number of red points for AT (29 out of 1000). Here, the adversarial loss corresponds to the classification margin which captures the confidence in correct classifications and is positive for correct predictions and negative for misclassifications. This further verifies that our proposed method improves the true  robustness and does not suffer from grading masking or obfuscation.

\subsection{Interpretability}
\begin{figure}[!tbp]
  \centering
  \subfloat[Saliency Maps.]{\includegraphics[width=0.4\textwidth]{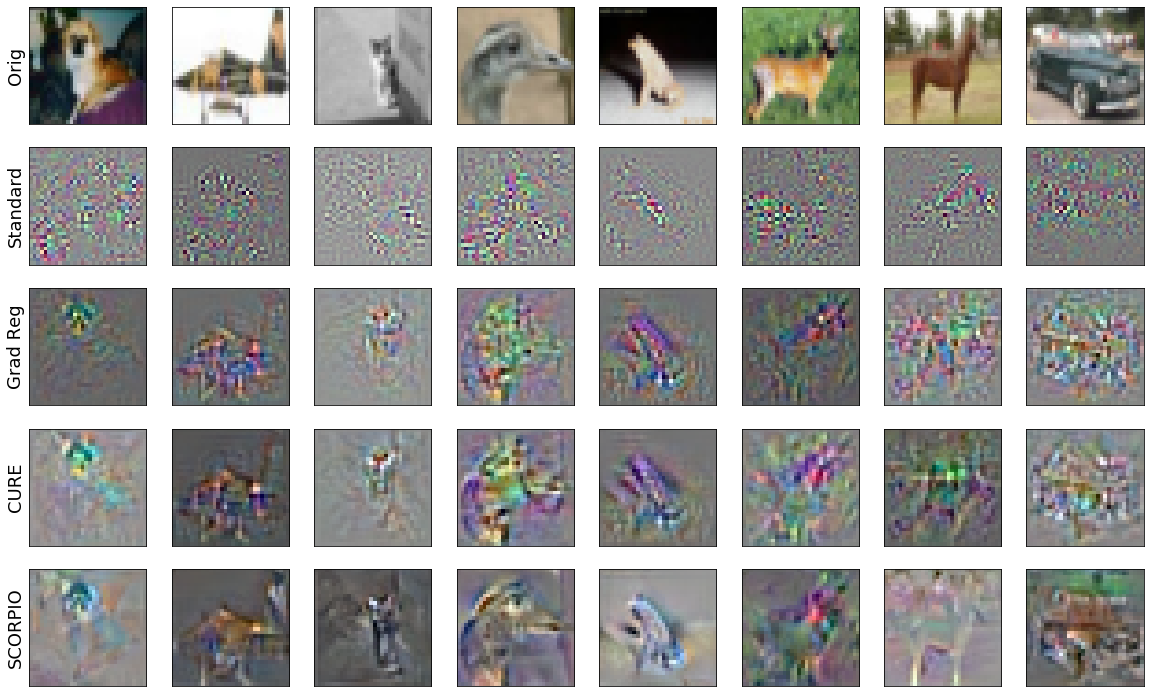}\label{fig:saliencymaps}}
  \hfill
  \subfloat[Adversarial Perturbations.]{\includegraphics[width=0.4\textwidth]{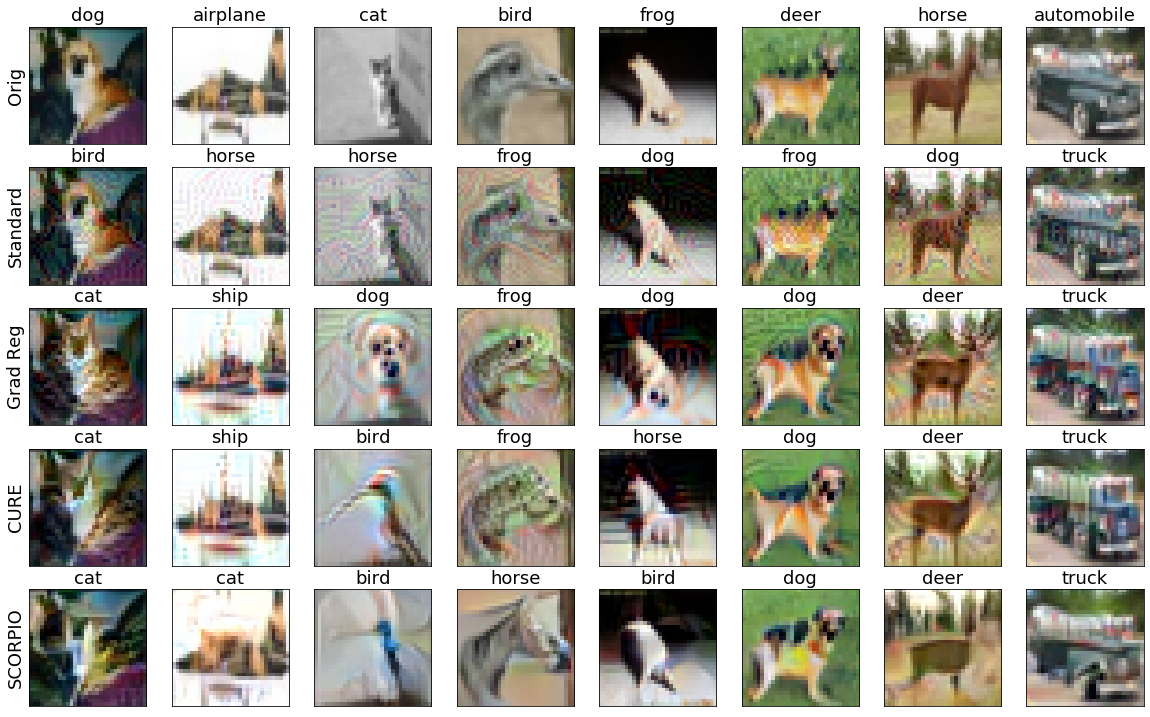}\label{fig:adversarial}}
  \caption{
  Saliency maps and PGD(20) untargeted adversarial attacks on the loss for sample examples from CIFAR-10 test set. We observe that our method SCORPIO achieves more realistic-looking adversarial perturbations and the quality of the saliency maps improves.
  }
\end{figure}

We compare the quality of the saliency maps generated with a variety of regularized networks, in addition to adversarial perturbations that can serve as counterfactuals for explanations. We note that standard networks produce very noisy saliency maps, as previously noted \cite{Etmann:2019}, and the higher level of robustness a network exhibits the more the network focuses on the object semantics as shown in Figure \ref{fig:saliencymaps}. The counterfactual images generated in Figure \ref{fig:adversarial} using (\ref{eq:L2attack}) align better with human perception for our method than for gradient and curvature regularized models.

\begin{figure}[!tbp]
  \centering
  \subfloat[$\delta_{max}$ Perturbations.]{\includegraphics[width=0.35\textwidth]{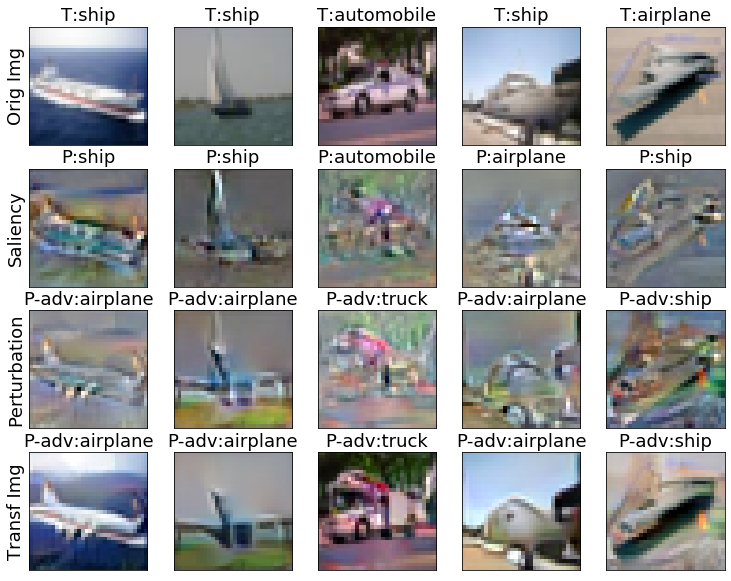}\label{fig:negs}}
  \hfill
  \subfloat[$\delta_{min}$ Perturbations.]{\includegraphics[width=0.35\textwidth]{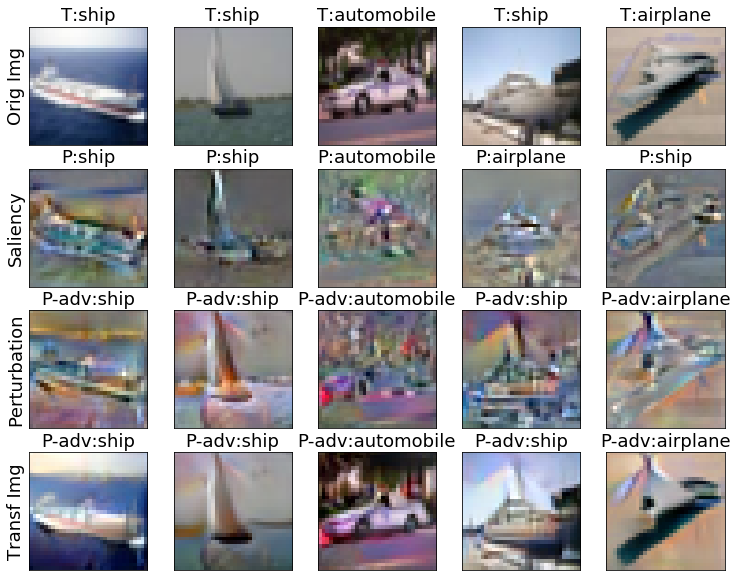}\label{fig:pos}}
  \caption{Contrastive explanations with SCORPIO on CIFAR-10 dataset. The perturbations were generated by solving (\ref{eq:L2opt}) and  (\ref{eq:L2attack}) via PGD with $\epsilon=4$ and 20 steps.}
\end{figure}

Figures \ref{fig:pos} and \ref{fig:negs} show example images from the CIFAR-10 test set, out of which the first three columns correspond to correct classifications and the last two columns correspond to misclassifications. Although saliency maps for our model focus more on the object of interest, identify areas in the image that most influence the predictions, and are a good sign of lack of gradient obfuscation \cite{qin:2019}, they cannot be directly used to reason about what features and how would need to change to correct a misclassification or how to justify a misclassification or a correct prediction.

To this end, we consider a framework inspired by contrastive explanations \cite{Dhurandhar:2019} and the robust feature manipulation by \cite{Tsipras2018:tradeoff}. Without loss of generality, we consider explaining decisions of an $L^2$-robust network\footnote{
This can be extended to $L^\infty$ as well, as we have noticed similar trends for robust models against $L^\infty$ adversaries.}. 
Our approach seeks to find \textit{pertinent negatives/positives} by optimizing over the perturbation variable $\delta$ that is used to explain the prediction. Pertinent negatives capture what is missing in the prediction, and pertinent positives refer to critical features that are present in the input examples.

Suppose we have $(x, y) \sim \mathcal{D}$, then we consider two contrastive
explanations defined by the optimizations
\begin{equation} \label{eq:L2attack}
    \delta_{\max}
        := \argmax_{\delta \in B_2(\epsilon)} l(x+\delta,y)
\end{equation}
and
\begin{equation} 
\label{eq:L2opt}
    \delta_{\min}
        :=\argmin_{\delta \in B_2(\epsilon)} l(x+\delta,y)
    ,
\end{equation}
we optimize the loss over a $L^2$ ball of radius $\epsilon$ to ensure that the modified example $x+\delta_m$ remains close to the original example $x$ in both cases. 

In the case of correct predictions $\delta_{\max}$ are the features within, or that could be added to, the original image which would flip the network's decision to a nearby class so are \emph{pertinent negative features} of the image. Whereas, $\delta_{\min}$ are the features which contribute to the correct prediction, making them the \emph{pertinent positive features}. In the case of incorrect predictions the roles are flipped. The roles are summarized below.

\begin{center}
\begin{tabular}{c|c|c}
    Pred. & $\delta_{\max}$ & $\delta_{\min}$  \\ \hline
    correct & PP & PN \\
    incorrect & PN & PP \\
\end{tabular}
\end{center}

For example consider the correct prediction in column 2 of Figure \ref{fig:negs} and \ref{fig:pos}. The network correctly predicts the image is of a ship and the
pertinent positive features, $\delta_{\min}$, are emphasizing the mast and box of the ship. The pertinent negative features, $\delta_{\max}$, show a nearby class for this image of a ship is airplane, but the features that need to be added to the original image in order to make it predict airplane are wings and shorter tail. Part of the explanation for this ship label is the \textit{absence} of these airplane features.

For an incorrect example consider column 4 of of Figure \ref{fig:negs} and \ref{fig:pos}. The network incorrectly predicts the image of a ship is an airplane. The $\delta_{\max}$ is now encoding the pertinent positive features and we see the perturbation emphasizes a portion at the bottom of the image which resembles landing gear. The pertinent negative features, now $\delta_{\min}$ suggest that the image lacks a clear distinction between the deck and hull and between the dock and water. As the transformed image $x+\delta_{\min}$ shows, the introduction of mast and mainsail features would correct this error. We remark that other elements in the image such as the sky and background are minimally perturbed.

These contrastive explanations coupled with adversarial learning have many potential uses such as: detect bias in training datasets, diagnose common errors made by a network, or distill concepts learned by the network, to name a few.

\section{Conclusion}

One of the most popular and effective training algorithms for robustness is adversarial training (AT), but is computationally expensive and does not incorporate curvature information into its attacks. We propose an approximate attack based on the quadratic-approximation to the adversarial risk that we show mathematically results in a worst case loss near the optimal adversarial loss. The attack is implemented via a Frank Wolfe iteration, which does not require projections or additional hyperparameter tuning, and a finite difference approximation that allows computing an effective quadratic approximate attack in less than half the number of forward-backward passes as AT reducing the training time. We incorporate these approximate attacks into a regularizer (SCORPIO) and demonstrate robustness near that of AT.
Further, SCORPIO suffers from less gradient obfuscation than AT and the quadratic-approximate attacks are shown to be as strong as PGD attacks against regularized models. 
Further, we use these robust models to develop an explanation framework based on pertinent positive/negative features.


\newpage

\small

\bibliographystyle{aaai21}
\bibliography{refs}

\section{Suplementary Material}

\subsection{Proof of Theorem \ref{thm:adv loss gap}}

\begin{proof}
Let $L_A$, $L_Q$, and $x$ be as stated in Theorem
\ref{thm:adv loss gap}. Further let $v_A$ and $v_Q$
be the corresponding adversarial and quadratic-approximate
attacks, respectively. Notice
$
    L_A - L_Q =
        \loss(x + v_A) - \loss(x) - (\loss(x + v_Q) - \loss(x))
        .
$
 Then by Taylor's Theorem there are degree 3 polynomials,
$R_{x,3}(v_A)$ and $R_{x,3}(v_Q)$ so that
\begin{align*}
    L_A - L_Q &=
        \loss(x + v_A) - \loss(x) - (\loss(x + v_Q) - \loss(x))
        \\
        &=
        Q(v_A) - Q(v_Q) + R_{x,3}(v_A) - R_{x,3}(v_Q)
        .
\end{align*}
Where we have suppressed the quadratic's dependence on $x$, i.e.,
$Q(v) := Q(v; x)$. Letting $g = \gradx \loss(x)$ and 
$H =  \hessx \loss(x)$
 we can write the difference of the quadratics as
\begin{align*}
    &Q(v_A) - Q(v_Q)
    \\
        &\quadd = 
            \inp{v_A}{g + \half H v_A}
            -
            \inp{v_Q}{g + \half H v_Q}
            \\
        &\quadd =
            \inp{v_A - v_Q}{g}
            +
            \inp{v_A}{\half H v_A}
            -
            \inp{v_Q}{\half H v_Q}
            \\
        &\quadd =
            \inp{v_A - v_Q}{g}
            +
            \inp{v_A - v_Q}{\half H v_A}
            \\
                &\quadd \quadd \quadd
            +
            \inp{v_Q}{\half H v_A}
            -
            \inp{v_Q}{\half H v_Q}
            \\
        &\quadd =
            \inp{v_A - v_Q}{g}
            +
            \inp{v_A - v_Q}{\half H( v_A +v_Q)}
\end{align*}
After taking the absolute value and applying the 
triangle inequality the gradient term is easily bounded with Holder, for
the remaining quadratic term
\begin{align*}
    |\inp{v_A - v_Q}{\half H( v_A +v_Q)}|
    &\leq
        \half 
        |v_A - v_Q|_p 
        \\
        &\quadd \times
        (|H v_Q|_q + |H v_A|_q)
        \\
    &\leq 
        {\epsilon^2}
        \nn H \nn_{p,q}
        .
\end{align*}
Therefore,
\begin{align*}
    | L_M - L_Q |
        &\leq
        2 \epsilon |g|_q
        +
        2 \epsilon^2 \nn H \nn_{p,q}
        +
        |R_{x,3}(v_A)|
        +
        |R_{x,3}(v_Q)|
        .
\end{align*}
The bound on the remainder terms is a result of Taylor's Theorem

\end{proof}

\subsection{Proof of Proposition \ref{lem:fw ascend}}
Assume the conditions of Proposition \ref{lem:fw ascend}

\begin{proof}
    Notice that $Q(g) \geq  \half g^t H g \geq 0$. Let $k \geq 1$ and write
    $v^{k+1} = v^k + \gamma^k( s^{k} - v^k)$. Then since
    $Q$ is quadratic it agrees with its second order Taylor expansion about $v^k$ and so
    \begin{align*}
        Q(v^{k+1})& - Q(v^k) =
                 \gamma^k \gradv Q(v^k) \cdot (s^{k} - v^k)
                \\
                &\quadd
                + \frac{(\gamma^k)^2}{2}(s^{k} - v^k)^T H (s^{k} - v^k)
                \\
            &\geq
                \gamma^k \gradv Q(v^k) \cdot (s^{k} - v^k)
                .
    \end{align*}
    The result follows by optimally of $s^k$.
\end{proof}

\subsection{Hyperparameters}
{\bf CIFAR-10.} For the $L^2$ experiments
the finite difference step and regularization strength, (h,r), used were: (1.15, 1.05) for FE(3) and FE(1),
(0.95, 0.999) and (1.25, 0.999) for CD(3) and CD(1). For the $L^\infty$ experiments: (1.05, 1.05) and (0.95, 1.05) for FE(3) and FE(1), (0.95, 1.15) for both CD(3) and CD(1). The learning rate was initialized at $0.01$ and decayed to $0.001$ after 50 epochs. Models
were trained for 60 epochs.

{\bf SVHN.} We tested only FE difference schemes with $N=3$ as these were the best performers in the CIRFAR-10 experiments. The models were trained for 40 epochs with an initial learning rate of $0.01$ that decayed by a factor of 10 every 8 epochs. The best model was chosen from all epochs.
For the $L^2$ experiments $h$ was $1.25$ and $r$ was $0.99$. For the $L^\infty$ the optimal $h$ was 1.5 and $r$ was 1.25.

\subsection{SVHN Results}
The trends seen in the CIFAR results are mirrored here
in the SVHN experiments. Table \ref{tab:L2 svhn attacks}
and \ref{tab:Linf svhn attacks} show that SCORPIO outperforms all geometric regularization methods and
is nearly as robust, and some cases more so, than AT. We
again see that untargeted attacks are stronger than targeted and that margin are somewhat stronger than loss attacks.

\begin{figure}
    \centering
    {\subfloat[$L^2$ adversarial PGD(10) attacks with loss / margin at $\epsilon=0.5$]{
    \begin{tabular}{l|ccccc}
    Method       & Clean & UL & TL & UM & TM  \\
    \hline
    Standard     & 97.0 & 21.2 & 45.6 & 20.7 & 45.8 \\
    Gradient Reg & 95.0 & 49.1 & 76.0 & 48.2 & 74.3 \\
    CURE         & 94.7 & 65.5 & 85.6 & 63.9 & 84.3 \\
    SCORPIO (FE,N=3) & 94.0 & 73.7 & 89.1 & 71.3 & 87.7 \\
    Adv Train (PGD-10)  & 95.3 & 72.9 & 89.0 & 71.7 & 88.2 \\
  \end{tabular}
  \label{tab:L2 svhn attacks}
    }}

    {\subfloat[$L^\infty$ adversarial PGD(10) attacks with loss / margin at $\epsilon=8/255$]{
      \begin{tabular}{l|ccccc}
    Method       & Clean & UL & TL & UM & TM  \\
    \hline
    Standard     & 97.0 & 2.4 & 18.8 & 2.4 & 20.5 \\
    Gradient Reg & 91.4 & 7.8 & 30.7 & 8.1 & 28.8 \\
    CURE         & 91.6 & 33.5 & 59.9 & 30.9 & 56.1 \\
    SCORPIO (FE,N=3) & 92.2 & 56.9 & 77.4 & 52.2 & 73.7 \\
    Adv Train (PGD-10) & 92.8 & 56.4 & 77.0 & 52.5 & 74.2 \\
  \end{tabular}
  \label{tab:Linf svhn attacks}
    }}
   
     \caption{Model accuracy on SVHN test set against various attacks. Our proposed regularizer, SCORPIO, outperforms prior gradient and curvature regularization methods achieves nearly the same level of robustness as AT in all cases.}
\end{figure}

\end{document}